\documentclass{article}
\usepackage{spconf,amsmath,graphicx}

\usepackage{enumitem}
\setlist{nosep, leftmargin=14pt}

\usepackage{color} 

\usepackage[normalem]{ulem}
\newcommand{\axa}[1]{\textcolor{blue}{#1}}
\newcommand{\axd}[1]{\sout{\textcolor{blue}{#1}}}

\usepackage{caption}
\usepackage{subcaption}
\usepackage[export]{adjustbox}

\def\x{{\mathbf x}}

\title{Self-Supervised Learning for Biological Sample Localization in 3D Tomographic Images}
\name{Author(s) Name(s)\thanks{Some author footnote.}}
\address{Author Affiliation(s)}
\name{Yaroslav Zharov\sthanks{Corresponding author: yaroslav.zharov@kit.edu}$^{\star \dagger}$ \qquad Alexey Ershov$^{\star}$ \qquad Tilo Baumbach$^{\star}$ \qquad Vincent Heuveline$^{\dagger}$}

\address{$^{\star}$ Laboratory for Applications of Synchrotron Radiation, Karlsruhe Institute of Technology, Germany\\
    $^{\dagger}$ Engineering Mathematics and Computing Lab, Heidelberg University, Germany}
\begin{document}
%
\maketitle
\begin{abstract}

In synchrotron-based Computed Tomography (CT) there is a trade-off between spatial resolution, field of view and speed of positioning and alignment of samples. 
The problem is even more prominent for high-throughput tomography--an automated setup, capable of scanning large batches of samples without human interaction. 
As a result, in many applications, only 20-30\% of the reconstructed volume contains the actual sample. 
Such data redundancy clutters the storage and increases processing time. 
Hence, an automated sample localization becomes an important practical problem.

In this work, we describe two self-supervised losses designed for biological CT.
We further demonstrate how to employ the uncertainty estimation for sample localization.
This approach shows the ability to localize a sample with less than 1.5\% relative error and reduce the used storage by a factor of four.
We also show that one of the proposed losses works reasonably well as a pre-training task for the semantic segmentation. 
\end{abstract}

\section{Introduction}
Computed tomography (CT) is a common imaging technique applied in biomedical imaging and material research, among others.
Advances in imaging technologies and availability of CT facilities lead to the widening gap between the capacity to produce huge amount of 3D data and the ability to analyse it.
This gap becomes even more pronounced in the serial, high-throughput tomography, capable of producing tens of volumes per hour.
This might be tackled using deep learning (DL) approaches, since they are proven to provide good results for a wide variety of image analysis problems.
However, DL models require an excessive amount of labelled data to achieve sufficient performance.

A variety of methods are developed to lower the amount of required training data or ease the markup process: weak supervision, transfer learning, self-training.
In this work we focus on the self-supervised learning (SSL).
It aims to develop a proxy task which does not require labelled data.
While the task itself could be unrelated to the final (downstream) task, it should force the model to produce useful representations.
For example, the model trained to colourize shades of grey photos and then fine-tuned for segmentation, performs better, than the model trained from scratch \cite{Larsson2017}.
Many methods aim at general-purpose representation learning, or also called data-driven priors \cite{Noroozi2016,Pathak2016,Jenni2018}.
However, there are also methods which learn representation based on the specific, domain-related knowledge, also called knowledge-driven priors \cite{Spitzer2018,Haghighi2021}.

Many SSL techniques are developed either for natural scene photography, or for medical data.
In biological CT, unlike in natural scenes, the imaging conditions and data acquisition protocols are usually fixed throughout an experiment.
However, biological samples, unlike medical ones, are rarely imaged in the exact same position.
This puts biological CT images somewhat in between of medical CT and natural scene photography.

In this paper we consider new methods that fall into the knowledge-driven category.
That is, to use our knowledge about the differences between the background and the sample structures, while avoiding relying on alignment between samples as much as possible.
We propose two self-supervised pretext tasks, and explore the usage of the learned representations for unsupervised sample localization and supervised semantic segmentation.

\section{Method}

We describe two self-supervised training procedures designed under an assumption that the object of interest is less homogenous than the background.
We denote them as a \emph{sorting loss} and a \emph{rotation loss}.
We also describe how to utilize \emph{uncertainty estimation} to localize samples in an unsupervised way.



\subsection{Pretext losses}

\begin{figure*}
  \centering
  \begin{subfigure}[b]{0.64\textwidth}
     \centering
     \includegraphics[width=\textwidth,keepaspectratio]{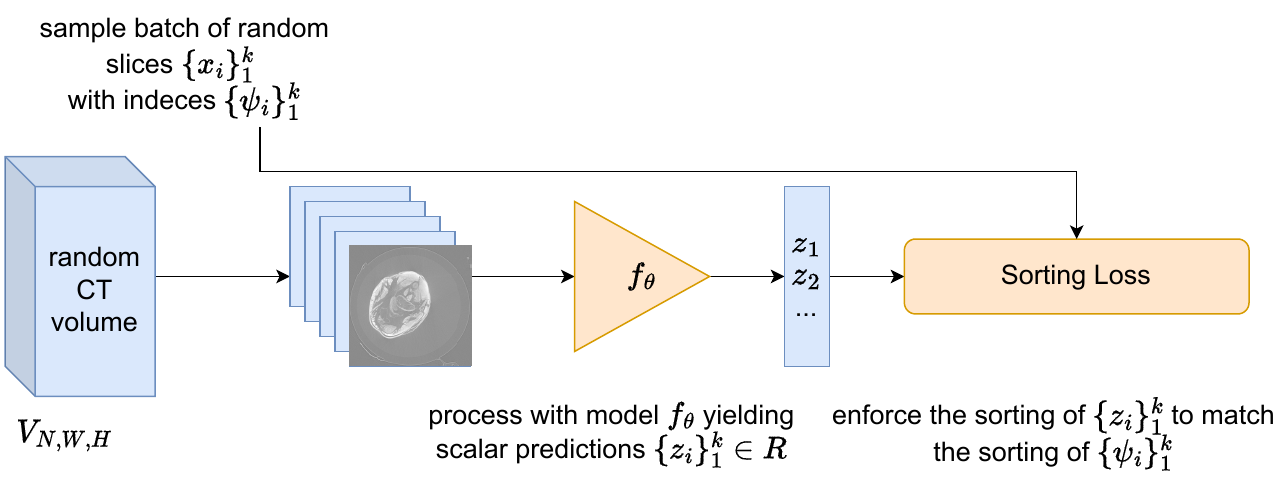}
      \caption{Formation and training of one batch. A single CT volume $V$ is sampled randomly for each batch. After position indices are sampled as $\psi_i \sim \mathcal{U}(1,N)$, we take slices as $x_i = V_{\psi_i}$. We used pairwise margin ranking loss as the training objective.}
      \label{fig:sortingloss_scheme}
 \end{subfigure}
 \hfill
\begin{subfigure}[b]{0.34\textwidth}
     \centering
     \includegraphics[width=\textwidth,keepaspectratio]{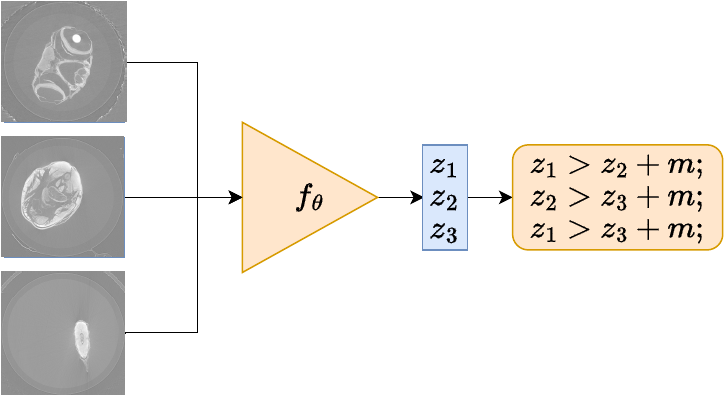}
      \caption{Intuitive example behind the \emph{sorting loss} calculation. Being fed with three slices of head, body, and tail, the model should predict three ascending values. The model for this should remember the outlook of the sample, and its structure.}
      \label{fig:sortingloss_example}
 \end{subfigure}
 \caption{Scheme (a) and intuitive example (b) of the sorting loss training procedure.}
 \label{fig:sortingloss}
\end{figure*}

The \emph{sorting loss} is inspired by the jigsaw puzzle task \cite{Noroozi2016}.
This loss requires the model to predict the order in a batch of slices, randomly taken along one axis.
For this loss to enforce the sample localization, the sample should be less permutation invariant (along the slicing axis) than the background.
Then, the easiest way to predict ordering is to learn features of the sample and their relative positions.
Due to typical absence of the sample alignment between different volumes in biological imaging, we propose to calculate the loss only between slices taken from the same volume.

During the training, we form a batch of size $k$ by randomly selecting one volume and then randomly sampling a set of slices $\{\x_i\}_1^k$ from this volume with uniformly sampled indices $\{\psi_i\}_1^k$.
We train the model $f_{\theta}$, which predicts one scalar value $z_i$ per input slice $x_i$.
We perform the optimization of the model with the pairwise \emph{margin ranking loss} \cite{Sculley2009}.
For each possible pair of predictions $z_i, z_j$ such that $\psi_i > \psi_j$, this loss enforces that $z_i \geq z_j + m$.
Where $m$ is the margin value to avoid collapse of the representation to small values.
Figure \ref{fig:sortingloss} shows the scheme of training (a) and intuition behind its work (b).

Without additional information, slice indices $\{\psi_i\}$ could be selected from the uniform distribution.
However, if the sample is known to have high margin to borders, it could be beneficial to use the generalized Gaussian distribution, to discourage sampling of empty slices.




As a control metric, we propose to use a mean displacement, which we define as a mean absolute distance between the predicted order and the true order of slices: $\frac{1}{k} \sum_i \vert \xi_i - \hat{\xi}_i \vert$, where $\xi = argsort([\psi_1,...,\psi_k]); \hat{\xi} = argsort([z_1,...,z_k])$.

\begin{figure*}
 \centering
 \includegraphics[width=0.8\textwidth,keepaspectratio]{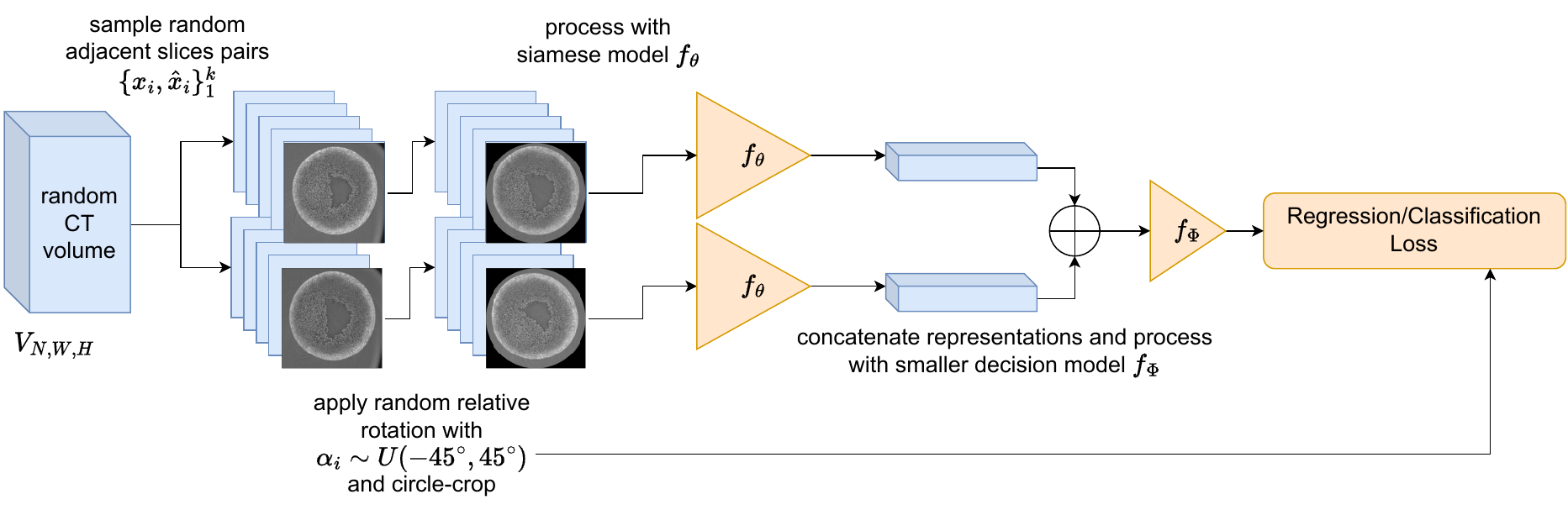}
  \caption{Formation and training procedure for one batch with rotation loss. The batch could be formed from several independent CT volumes. After sampling indices $\psi_i \sim \mathcal{U}(1, N)$ and $\hat{\psi}_i \sim \mathcal{U}(\psi_i-l, \psi_i+l)$, the slices are taken as $x_i = V_{\psi_i}$ and $\hat{x}_i = V_{\hat{\psi}_i}$. The loss is calculated between the angle predicted by $f_{\Phi}$ and $\alpha_i$ used for different augmentation.}
  \label{fig:rotationloss}
\end{figure*}

The \emph{rotation loss} is based on the rotation-prediction loss for photography images \cite{Gidaris2018}.
We adapt it to the specifics of the biological CT images.
In the original work, the authors proposed to predict the rotation angle of a randomly rotated image.
Since, in biological CT, samples are typically randomly oriented, there is no common reference orientation.
Hence, we train the model to predict the rotation angle between two randomly rotated slices.
To avoid overfitting to the local imaging and reconstruction noises, we use two adjacent slices instead of rotating copies of the same slice, which could be seen as a natural augmentation.
To avoid overfitting to the rotation artefacts (appearing for angles other than multiplies of $90^{\circ}$), we crop each slice with an incircled mask of maximal diameter.

For this loss to force the model to learn object-related features, an object should be less rotationally invariant than the background at least along one axis.
For example, on the left, axial view on Figure \ref{embryo_box_sample}, the sample is less rotationally invariant than the background (which appears as a circle), and on the right, the same sample is more rotationally invariant than the background (which is asymmetrical).

We pass each pair of slices through a siamese model $f_{\theta}$ \cite{Chicco2021}.
Then, we use an additional small model $f_{\Phi}$, to predict the rotation angle, based on concatenated representations from two branches.
We observed, that for practical purposes it's easier to train the model with the classification loss instead of regression. 
In this work we use a negative log-likelihood, and leave the study of different losses for the future research.
The data processing and training procedures are depicted on the Figure \ref{fig:rotationloss}.


\subsection{Uncertainty Estimation for Unsupervised Localization}

To employ the learned representations for the unsupervised localization we assume, that, given the good generalization during training, the model have learned to rely on the features of a sample rather than of a background.
This assumption stays in the line with the assumptions that were made earlier for the pre-training losses. 

For the \emph{sorting loss}, we assume that since the model have learned to predict the position of the slice using the sample structures, dropping out those important regions will lead to the higher uncertainty.
We obtain embeddings of the superpixels for each slice by removing from the model the Global Average Pooling layer, and the last linear layer.
By letting the model work with train-time dropout, we sample from the distributions of the superpixel embeddings.
We find the region of interest, by thresholding the variation of these distributions.

For the sample localization after the \emph{rotation loss}, we replace the \emph{3D localization} task with a series of \emph{2D detection} tasks, for each slice along each orthogonal axis.
A slice is denoted as containing an object if the uncertainty of the prediction is low.
To estimate uncertainty during the test time, we process the non-rotated copies of the same slice.
Since the classification model outputs probabilities for each class, we use $1 - p_0(x_i, \hat{x}_i)$ as the uncertainty estimation, where $p_0$ is the probability predicted by the model for the $0^\circ$ rotation angle.

We selected the simplest ways to estimate uncertainty for each case.
We leave the study of different uncertainty estimation methods for the future research.

\section{Experiments}
We investigate the applicability and the properties of the proposed approaches with the following experiments.
To qualitatively explore the localization using the \emph{sorting loss}, we use a CT dataset of the \textit{Medaka} fish \cite{Weinhardt2018}.
The dataset contains 250 volumes of average size $6100 \times 2000 \times 2000$ pixels, with slightly varying size along the first axis.
We also quantitatively measure the effect of the pre-training with the \emph{sorting loss} for semantic segmentation on the same dataset.
For this, we obtained segmentation markups of the visual system of the fish for 15 volumes.

Further, we quantitatively study the localization with the \emph{rotation loss}.
We use a CT dataset of \textit{Xenopus Laevis} embryos \cite{Moosmann2013}.
It consists of 34 volumes of size $2016 \times 2016 \times 2016$ pixels.
For the training and processing purposes, we downscaled the volumes to $504 \times 504 \times 504$ pixels.
We setted up the bounding box labels manually to check the ability of the model to perform localization.


\subsection{\textit{Medaka} Fish Localization}
From the whole dataset, we selected 50 volumes and randomly split them in 3\axa{:}\axd{-}1 proportion to a train and a validation set.
We trained a ResNet-18 model with the \emph{sorting loss}, with the margin set to $m = 0.2$.
Since the mean displacement reached $0.42$ on the test set and $0.26$ on the train set for the batch size of $12$, we conclude that the estimator provides reasonably good ordering.
To define the bounding box, we used the largest connected region of the 3\% super-pixels with the highest variance of the embedding.

\begin{figure}[h]
    \begin{subfigure}{0.45\textwidth}
        \centering
        \adjincludegraphics[width=\textwidth,keepaspectratio,trim={0 0 {.5\width} 0}, clip]{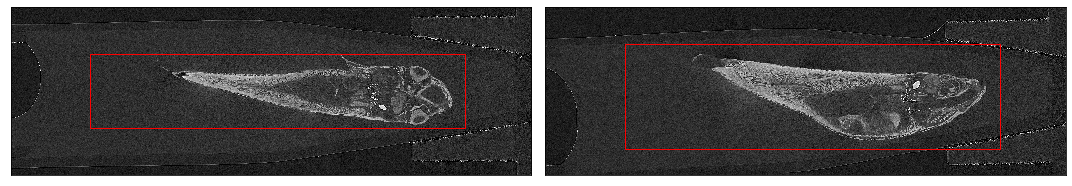}
    \end{subfigure}
    \begin{subfigure}{0.45\textwidth}
        \centering
        \includegraphics[width=\textwidth,height=0.7\textheight,keepaspectratio]{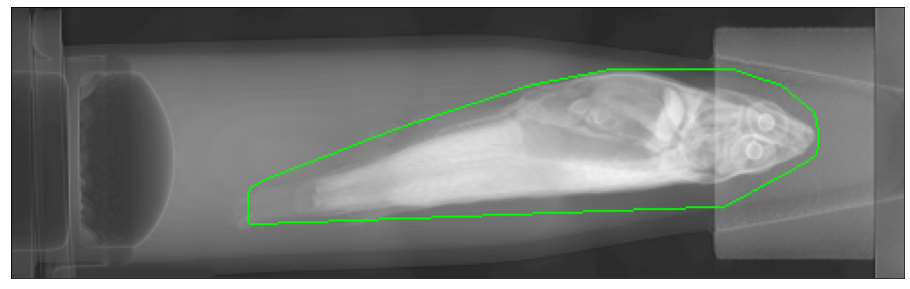}
    \end{subfigure}
    \caption{Examples of localization for the \textit{Medaka} fish dataset done with the \emph{sorting loss}.
            \textbf{Top}: slice through the volume with the predicted 3D bounding box.
            \textbf{Bottom}: projection of the volume with contour of the convex hull surrounding the predicted region of interest.
            It can be observed, that the method does not rely solely on the grey values, as the bright container is excluded from the bounding box, while the less intensive tail is included.}
    \label{fish_sample}
\end{figure}

In Figure \ref{fish_sample} we present examples of the localization done for a \textit{Medaka} fish sample.
The quality of the estimated bounding box results was sufficient to drive the cropping procedure without losing meaningful information. 
It reduced the size of the volume by $74\%$ per volume on average, i.e. from $24$ Gb to $5$ Gb.

\subsection{\emph{Medaka} Fish Segmentation}

To investigate if the proposed loss can act as a pre-training, we compared it with the SimCLR and MoCoV3.
We prepared four ResNet-18 snapshots.
Three pretrained with the \emph{sorting loss}, SimCLR, and MoCoV3, and one randomly initialized \cite{Chen2020b, Chen2021}.
We then trained four DeepLabV3+ models, with those snapshots used as the encoder initialization.
From the results presented in the Table \ref{tab:medaka-small} we conclude that the proposed method works on par with baselines in the low data regime, and does not harm the model performance in the full data regime.
Possible harmful results of pre-training were described previously \cite{Zoph2020}.

\begin{table}[]
    \centering
    \begin{tabular}{r||c|c|c|c}
        & \multicolumn{4}{c}{pre-training}  \\
        data \% & None & SimCLR & MoCoV3 & \emph{sorting loss} \\ \hline \hline
        1 &  $63.5$ & $66.6$ & $\mathbf{67.9}$ & $67.4$ \\ \hline
        100 & $74.8$  & $72.7$ & $74.5$ & $\mathbf{75.0}$
    \end{tabular}
    \caption{IoU of Medaka visual system segmentation, depending on the pre-training loss and amount of supervised dataset used. The median values are estimated on 5 independent runs.}
    \label{tab:medaka-small}
\end{table}

To investigate the ability to use the pre-training of large models on large data, we performed pre-training of the ResNet-152 model on the full dataset.
We found out, that even while fine-tuning U-Net (a model with a large decoder), we were able to improve IoU by 2.3\% compared to training from scratch.

Additionally, we noted that the aggressive augmentation can further increase quality for the \emph{sorting loss} pre-training, while decreasing it for the SimCLR.
These results agree well with the recent findings \cite{Wang2021}.
Furthermore, the proposed method doesn't use double sampling or dual models, which leads to less computational overhead per image.

\subsection{\textit{Xenopus Laevis} Embryo Localization}

We trained a ResNet-50 model with the classification version of the rotation loss.
The set of possible rotation angles was from $-90^\circ$ to $90^\circ$ with a step of $30^\circ$.
For each reference slice, we sampled the twin-slice within $100$ slices range.

\begin{figure}
  \centering
  \includegraphics[width=.8\linewidth]{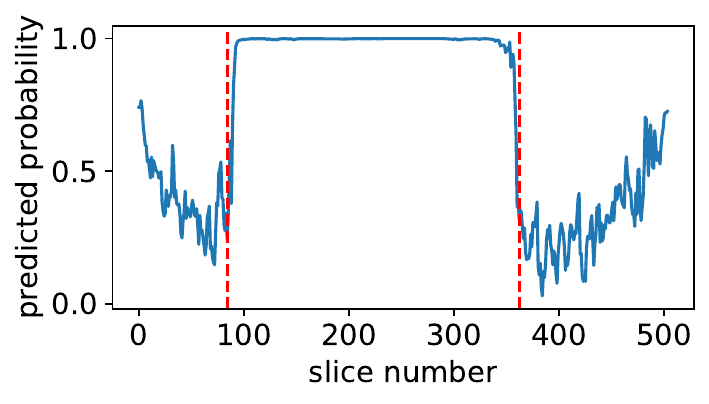}
  \captionof{figure}{
  Probabilities of the correct rotation class for all slices along one slicing axis of \textit{Xenopus Laevis} embryo.
  Ground truth position of the embryo shown with dashed red lines.}
  \label{embryo_prob_sample}
\end{figure}

As a sanity check of the uncertainty estimation approach, we provide a plot of the certainty along a single axis (Figure \ref{embryo_prob_sample}).
The area predicted to belong to the embryo have consistently higher probability than the area related to the background.
We conclude that the model was trained to use intrinsic features of the embryo and not of the background to solve the rotation estimation task.

\begin{figure}
\centering
  \centering
  \includegraphics[width=.8\linewidth]{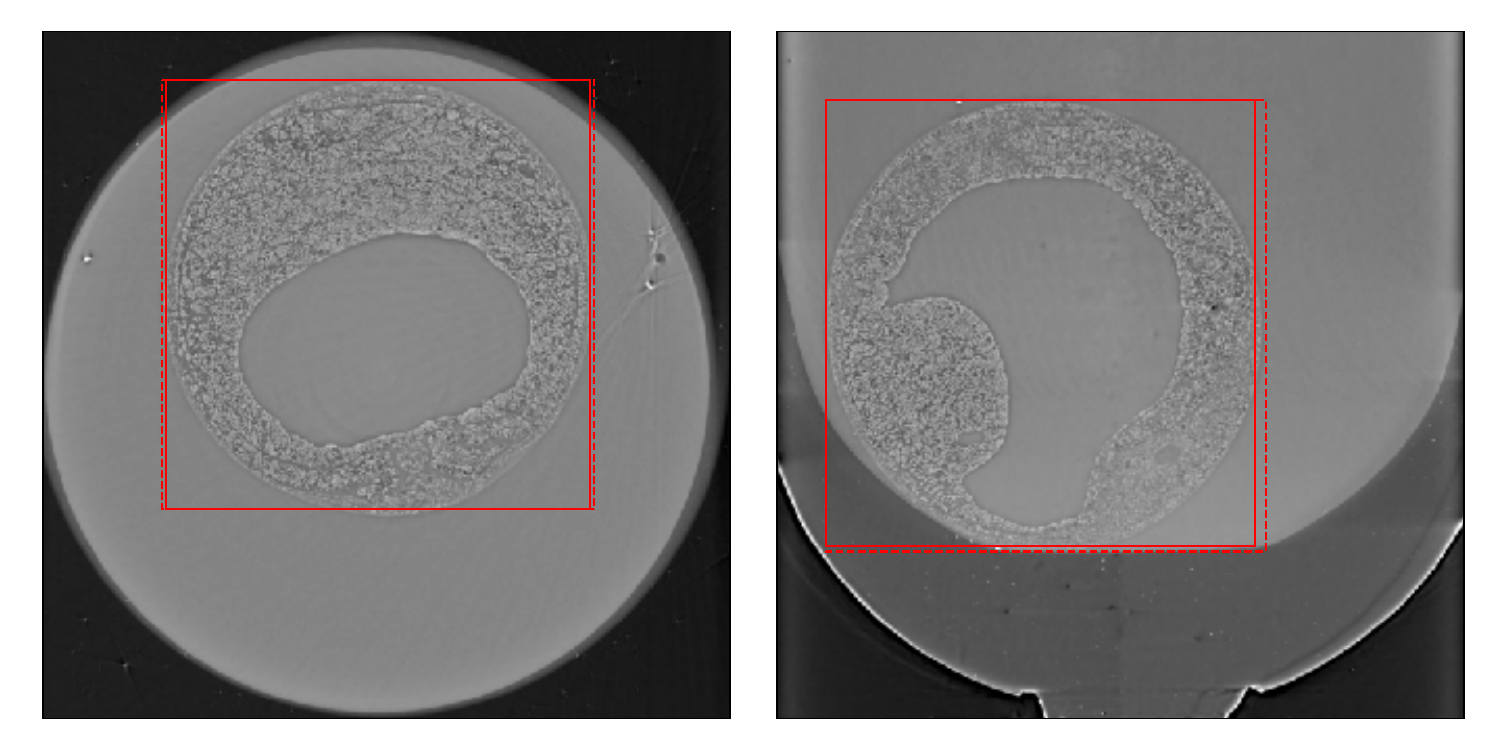}
  \captionof{figure}{Predicted 3D bounding box for the \textit{Xenopus Laevis} embryo depicted with a solid line and the ground truth depicted with a dashed line. Left: Axial view. Right: Coronal view.}
  \label{embryo_box_sample}
\end{figure}

The Mean Absolute Error (MAE) of bounding box coordinates is $3.5$ pixels.
As a baseline, we took thresholding by the pixel value, and after a careful threshold selection (with supervision available on the whole set), the best achievable MAE was $12.8$
We also provide a qualitative bounding box comparison in Figure \ref{embryo_box_sample}.


\section{Conclusion \& Further work}
We presented two approaches of self-supervised learning designed specifically for the \textit{volumetric} X-ray CT biological data.
We described two pretext tasks and respective loss functions, and a way to employ uncertainty estimation to transfer the trained models to a downstream task of unsupervised object localization.

Presented approaches show the ability to localize sample with $1.2\%$ relative error, and practically allow shrinking the used storage size by $74\%$ via cropping of the original 3D volume.
We also demonstrated that \emph{sorting loss} can work as a pre-training for the downstream task of supervised segmentation. 
It provides results that are slightly better than the baseline, while taking less resources for training.

As a future work, we plan to employ the proposed method to replace an expert-driven bounding box selection for the weak supervision for semantic segmentation \cite{NIPS2019_8885}, as well as to use those methods as a pre-training.
The code and the relevant datasets will be made publicly available.

\section{Compliance with Ethical Standards}
The data used in this paper was collected for other projects. 
No additional experiments were conducted.
The article will be updated upon the dataset publication.

\section{Acknowledgments}
\label{sec:acknowledgments}
We acknowledge the support by the projects CODE-VITA (BMBF; 05K2016) and HIGH-LIFE (BMBF; 05K2019).
We gratefully acknowledge the data storage service SDS@hd supported by the Ministry of Science, Research and the Arts Baden-Württemberg (MWK) and the German Research Foundation (DFG) through grant INST 35/1314-1 FUGG and INST 35/1503-1 FUGG.
We acknowledge the KIT for provision of instruments at the Karlsruhe Research Accelerator (KARA) and thank the personnel of imaging beamlines.
We thank Sabine Bremer for the provided \textit{Medaka} fish samples and Janes Odar for the provided \textit{Xenopus Laevis} samples. 

\bibliographystyle{IEEEbib}
\bibliography{references}

\end{document}